\documentclass[conference]{IEEEtran}
\IEEEoverridecommandlockouts
\usepackage{cite}
\usepackage{amsmath,amssymb,amsfonts}
\usepackage{algorithmic}
\usepackage{graphicx}
\usepackage{textcomp}
\usepackage{xcolor}
\usepackage{hyperref}
\usepackage{booktabs}
\usepackage{xcolor}

\def\ie{\emph{i.e.}}

\def\BibTeX{{\rm B\kern-.05em{\sc i\kern-.025em b}\kern-.08em
    T\kern-.1667em\lower.7ex\hbox{E}\kern-.125emX}}
\begin{document}


\title{EgoSpot: Egocentric Multimodal Control for Hands-Free Mobile Manipulation}

\author{
Ganlin Zhang$^{1,3*\dagger}$,
Deheng Zhang$^{2,3*\dagger}$
Longteng Duan$^{3*}$, 
Guo Han$^{3*}$,\\ 
Yuqian Fu$^{4}$, 
Danda Pani Paudel$^{2}$, 
Luc Van Gool$^{2}$, 
Eric Vollenweider$^{5}$ \\
{\small $^{1}$Technical University of Munich}
{\small $^{2}$INSAIT, Sofia University “St. Kliment Ohridski”}
{\small $^{3}$ETH Zurich}
{\small $^{4}$SJTU}
{\small $^{5}$Microsoft}\\
{\small *Equal contribution}
{\small \textdagger Work done while at ETH Zurich}
}

\maketitle

\begin{figure}[t]
    \centering
    \includegraphics[width=0.5\textwidth]{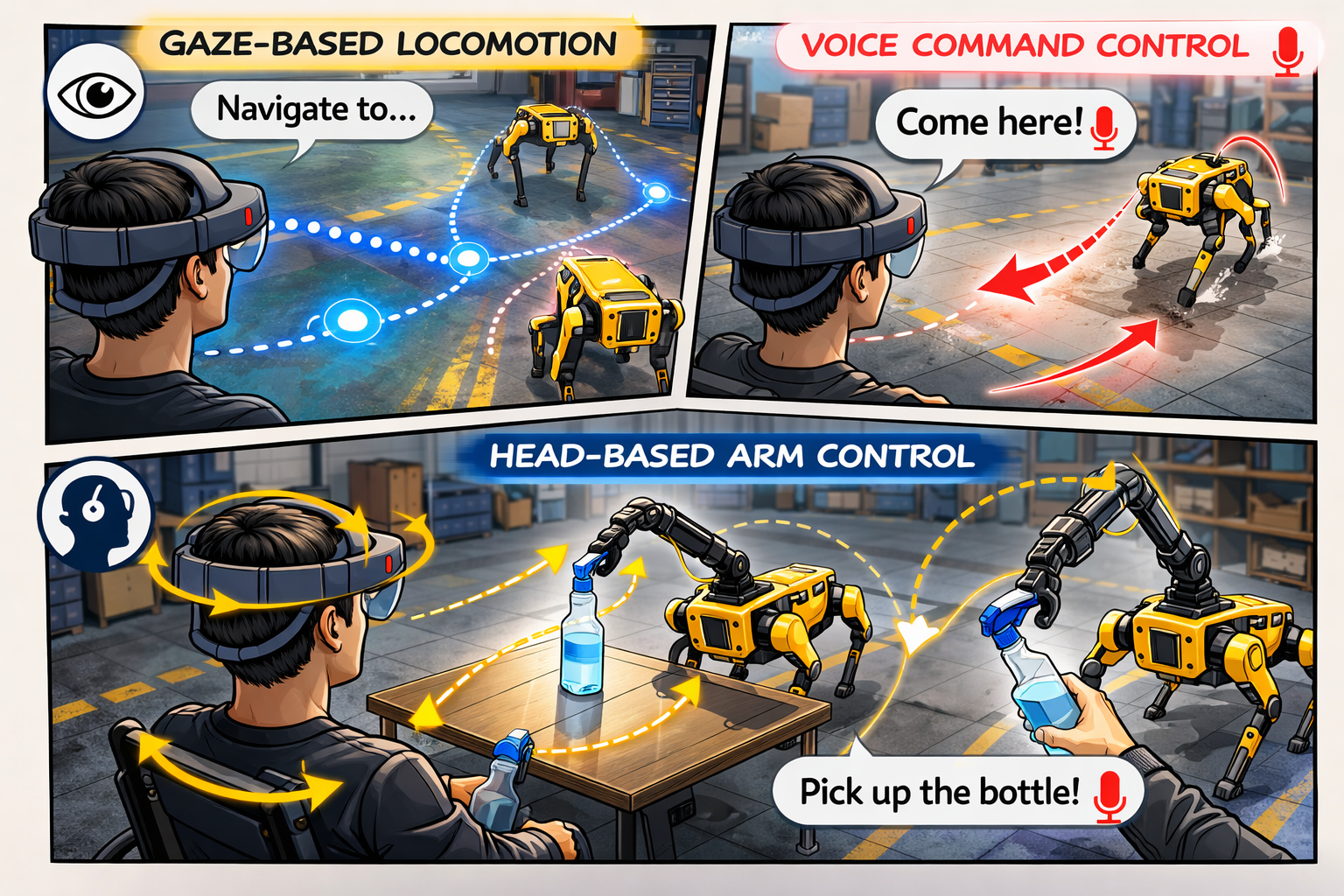}
    \caption{
    Overview of our accessible multimodal control for mobile manipulation.
    }
    \label{fig:teaser}
    \vspace{-2em}
\end{figure}
\begin{abstract}
We propose a novel hands-free control framework for the Boston Dynamics Spot robot using the Microsoft HoloLens 2 mixed-reality headset. Enabling accessible robot control is critical for allowing individuals with physical disabilities to benefit from robotic assistance in daily activities, teleoperation, and remote interaction tasks. However, most existing robot control interfaces rely on manual input devices such as joysticks or handheld controllers, which can be difficult or impossible for users with limited motor capabilities. To address this limitation, we develop an intuitive multimodal control system that leverages egocentric sensing from a wearable device. Our system integrates multiple control signals—including eye gaze, head gestures, and voice commands—to enable hands-free interaction. These signals are fused to support real-time control of both robot locomotion and arm manipulation. Experimental results show that our approach achieves performance comparable to traditional joystick-based control in terms of task completion time and user experience, while significantly improving accessibility and naturalness of interaction. Our results highlight the potential of egocentric multimodal interfaces to make mobile manipulation robots more inclusive and usable for a broader population. A demonstration of the system is available on our \href{https://ganlinzhang.xyz/Holo-Spot-Page/index.html}{project webpage}.

\end{abstract}


\section{Introduction}

In recent decades, rapid technological advances in robotic manipulation and control~\cite{he2025sequential, wang2025scoop, wang2026ocraobjectcentriclearning3d} have reshaped many aspects of daily life. At the same time, increasing awareness of accessibility, diversity, and social inclusion has encouraged the development of technologies that empower individuals with physical disabilities and enable them to participate more fully in society. Access to technology is not only a matter of convenience, but also of equity, independence, and dignity. In this context, assistive technologies~\cite{nanavati2023assistive,sarsenbayeva2022motor} and accessible human--computer interfaces play a critical role by allowing users with limited motor capabilities to interact with digital systems and robotic platforms. Such systems have the potential to reduce barriers in daily activities, expand opportunities for communication and mobility, and support greater participation in education, work, and social life.

In particular, accessible robot control is an important yet underexplored problem. Mobile manipulation robots are increasingly being deployed in real-world scenarios such as inspection~\cite{pearson2024substation}, remote operation~\cite{dass2024telemoma}, search and rescue~\cite{schwarz2017momaro}, and assistive services~\cite{park2020feeding}. However, most existing robot control systems rely on handheld devices such as joysticks, keyboards, or controllers that require precise manual operation, posing significant barriers for users with upper-limb impairments.

More fundamentally, there exists a mismatch between human perception and robot control interfaces. Humans naturally perceive and act from an egocentric perspective~\cite{fathi2011understanding, plizzari2024outlook, fu2025objectrelator, li2025egocross, zhang2025egonight}, leveraging first-person signals such as gaze, head motion, and speech to guide actions in a continuous and embodied manner. In contrast, most robot control systems remain device-centric and detached from this egocentric perception. This gap motivates the need for egocentric interaction frameworks that directly translate first-person sensing into robot control, enabling intuitive and hands-free operation.


{In this work, we present \textbf{EgoSpot}, an egocentric control framework for accessible mobile manipulation using the Boston Dynamics Spot robot and a Microsoft HoloLens~2 headset. Building upon the egocentric interaction paradigm, EgoSpot translates first-person sensing signals into actionable robot commands, enabling hands-free control without reliance on manual input devices.

Specifically, our framework integrates multiple egocentric modalities—including eye gaze, head motion, and voice commands—within a unified control pipeline to support real-time robot locomotion and arm manipulation. To ensure consistent interaction between the user and the robot, we align the egocentric reference frame of the mixed-reality interface with the robot coordinate system through an inverse transformation method based on Azure Spatial Anchors. This alignment allows users to directly control robot motion and manipulation from their first-person perspective, effectively grounding robot actions in egocentric perception.

\begin{enumerate}
\item We introduce an \textbf{egocentric multimodal interaction paradigm} for accessible robot control, which enables hands-free operation of mobile manipulation robots by directly leveraging first-person sensing signals, including gaze, head motion, and speech.

\item We propose a method to \textbf{align the egocentric reference frame with the robot coordinate system}, allowing control commands to be consistently grounded in the user’s first-person perspective.


\item Through user studies, we show that the proposed system provides effectiveness and usability comparable to conventional hand-based control interfaces.



\end{enumerate}

\section{Related Works} 
\label{sec:related_works}
\subsection{Accessible Robot Manipulation}
Mixed reality has recently emerged as an important research direction for robot control and task execution. Prior studies have explored the use of mixed reality devices for controlling robotic arms~\cite{gadre2019end,neves2018application,ostanin2018interactive}, while~\cite{wu2020mixed} applies mixed reality to mobile robot path planning. Despite their promise, these systems are generally not designed with accessibility in mind, as most require hand gestures as a primary interaction modality and are therefore not well suited for users with amputations or limited hand mobility. To enhance interaction accuracy, several works have incorporated additional modalities. For example,~\cite{kyto2018pinpointing} combines hand gestures with eye tracking for more precise object selection, and~\cite{krupke2018comparison} uses head-based pointing or gesture input together with speech to control a robotic arm. However, these approaches remain focused mainly on manipulation tasks rather than mobile robot navigation, and none fully eliminates the need for hand operation.
\subsection{Egocentric Perception for Robot Learning}
In parallel, recent advances in robot learning have increasingly leveraged
human demonstrations, especially from egocentric and first-person viewpoints.
For example,~\cite{zhou2025yoto} extracts bimanual pose and motion patterns
from binocular hand demonstrations for visuomotor imitation, while
~\cite{yang2025egovla} trains vision--language--action models from egocentric
human videos and retargets human actions to robots through inverse kinematics.
Similarly,~\cite{zhang2025robowheel} proposes a data engine for reconstructing
hand--object interactions from monocular videos and retargeting the resulting
trajectories to robots with different embodiments. Furthermore,
~\cite{song2025mitty} introduces a diffusion-based framework that translates
egocentric human demonstrations into robot-execution videos in an end-to-end
manner, and~\cite{yao2025think} explores primitive prompt learning for lifelong
robot manipulation. Beyond these works, recent research further strengthens the
connection between egocentric perception and robot learning. EgoMI studies how
active head motion and hand--eye coordination in first-person demonstrations can
be modeled for whole-body manipulation, highlighting the importance of
egocentric viewpoint dynamics when transferring human behavior to
robots~\cite{yu2025egomi}. More recently, HoMMI demonstrates that egocentric
sensing can support robot-free collection of whole-body mobile manipulation
data, enabling scalable learning of navigation and manipulation skills from
human demonstrations~\cite{xu2026hommi}. Taken
together, these works suggest that egocentric sensing provides a natural
foundation for intuitive, context-aware, and transferable robot control.
\section{System Design} 
\label{sec:system_design}
\subsection{Overall Framework}
\begin{figure}[htbp]
    \centering
    \includegraphics[width=\linewidth]{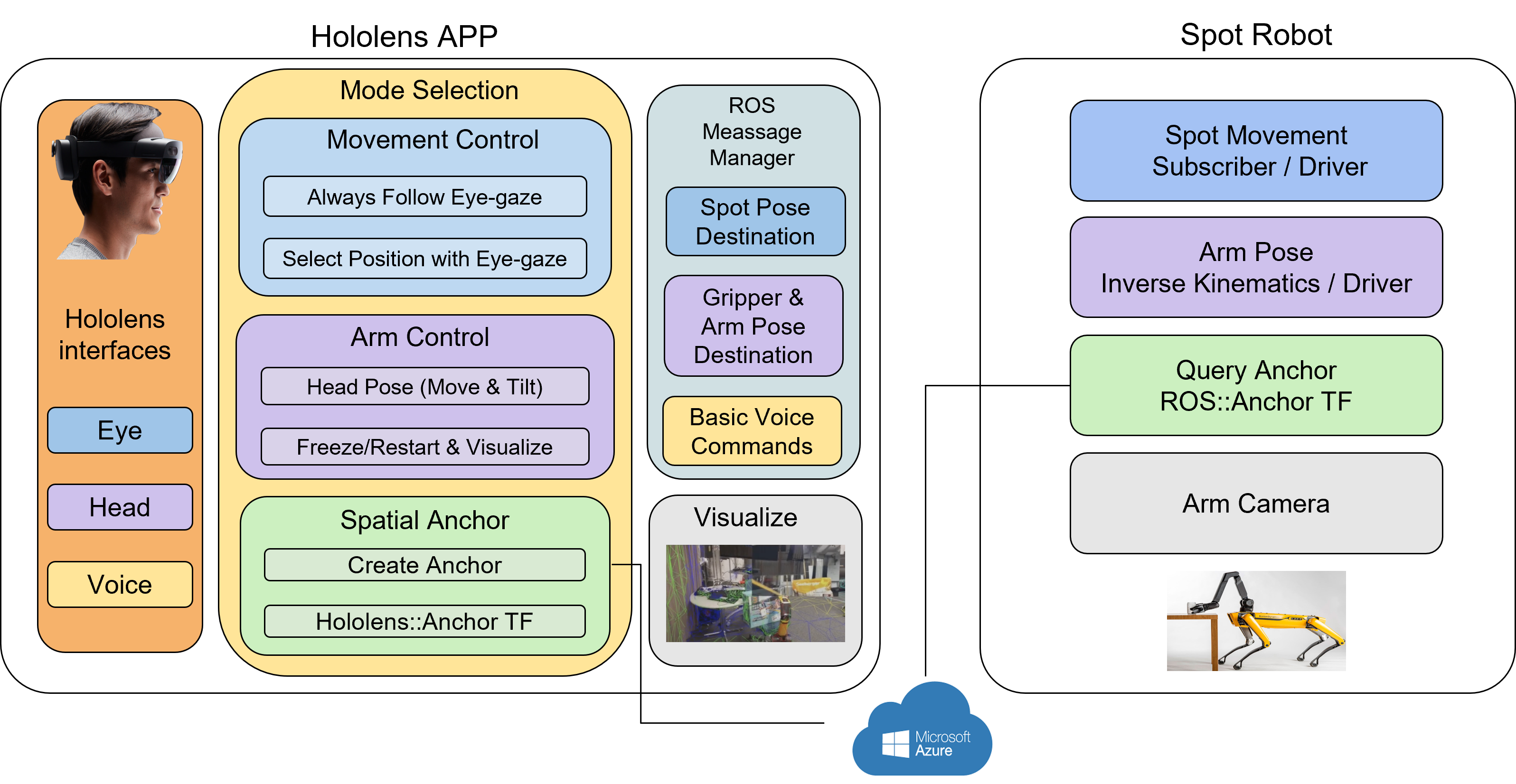}
    \caption{Overall system architecture. The HoloLens 2 app (left) runs a Unity-based mixed-reality interface controlled by eye gaze, head motion, and voice commands. Azure Spatial Anchors provide co-localization between the headset and the Spot robot. Control messages are sent via ROS over TCP to the robot, which executes locomotion (Follow/Select modes) or arm manipulation (Arm mode) based on the active mode and user inputs.}
    \label{fig:system}
\end{figure}

Our system architecture is illustrated in Fig.~\ref{fig:system}. The overall system consists of two main components: a HoloLens application developed in Unity and a ROS-based control stack running on the Spot robot. Azure Spatial Anchors are used to achieve co-localization between the HoloLens and the robot, enabling both systems to share a common spatial reference frame. To support users with upper-limb amputations, the application is designed to be fully hands-free and relies on three input modalities: eye gaze, head motion, and voice commands.

Voice commands are used both for issuing basic robot instructions, such as \textit{sit} and \textit{stand}, and for switching between different control modes, including robot locomotion, arm and gripper control, and spatial anchor creation. Within each mode, users interact with the robot through eye gaze and head motion. During operation, the HoloLens continuously publishes ROS messages to the Spot robot, including information such as target body positions and desired arm poses. The robot continuously listens for these messages, queries the spatial anchor when needed, performs the required coordinate transformations, and executes the corresponding actions. Overall, the system functionality can be divided into two categories: robot body control and robot arm control. All functions are initiated and managed through voice commands, allowing users to enter a specific mode and activate or terminate it as needed.

\textbf{Basic Voice Commands.} The system supports more than ten basic voice commands that allow users to perform fundamental robot actions. These include \textit{sit}, \textit{stand}, \textit{power on}, \textit{power off}, \textit{claim}, \textit{release}, \textit{self right}, \textit{roll over left}, \textit{roll over right}, \textit{spin left}, and \textit{spin right}. Their functions are intuitive: once the robot receives a command, it immediately executes the corresponding action. One particularly useful command is \textit{come here}, which instructs the robot to navigate to the current position of the HoloLens wearer.

\textbf{Follow Mode.} This mode is activated by the voice command \textit{follow mode}. Once activated, the Spot robot follows the user’s eye gaze direction. To provide clear visual feedback, a sphere cursor is displayed to indicate the current gaze target in the environment.

\textbf{Select Mode.} This mode is activated by saying \textit{select mode}. In this mode, users first specify a target position by saying \textit{select item}. A white cube is then displayed at the selected location as visual confirmation. When select mode is activated, the Spot robot navigates directly to the chosen target position. Users can pause the robot at any time by saying \textit{terminate}, and resume movement toward the same target by saying \textit{activate}. At any given time, only one target position can be selected.

\textbf{Arm Mode.} This mode is selected by saying \textit{arm mode}. After entering the mode, users can say \textit{activate} to begin controlling the robot arm, which then follows the user’s head movements. A live video stream from the robot hand camera can be toggled on or off using the commands \textit{visualize on} and \textit{visualize off}; this video feed appears in the upper-right corner of the display as shown in Fig.~\ref{fig:video_livestream}. Users can freeze the arm at its current pose by saying \textit{terminate}. They may then move to a different physical position and reactivate the mode, after which the arm resumes control from the previously fixed pose. This interaction design simplifies arm manipulation and makes it easier to perform precise adjustments. In addition, users can say \textit{rotate hand} and \textit{stop rotate hand} to enable or disable gripper rotation. When rotation is enabled, tilting the head to the left or right causes the gripper to rotate in the corresponding direction. Finally, users can say \textit{grasp} to toggle the gripper state, using the same command to alternately open and close it.

\subsection{Hololens}
\textbf{Communication.} On the HoloLens side, communication between Unity~\cite{unity} and ROS~\cite{ros} is implemented using the ROS--TCP Connector package~\cite{ROS_TCP}. For modes requiring continuous updates, such as Follow Mode and Arm Mode, messages are published at a fixed frequency instead of every frame. While all modes use the same message type, they publish to different topics and encode different information. All topics are registered at application startup.

\textbf{Mode Switching.} Because the system supports multiple control modes and a conventional GUI is not appropriate for our target users, assigning a separate voice command to every action would lead to an overly complex command set. We therefore reuse common voice commands across modes while restricting certain commands to specific modes. For instance, \textit{grasp}, \textit{rotate hand}, and \textit{stop rotate hand} are only valid in Arm Mode, whereas \textit{select item} and \textit{delete selection} are only available in Select Mode. To manage this behavior, we adopt the state pattern~\cite{dyson1996state}, as illustrated in Fig.~\ref{fig:state_pattern}. An \textit{OperationMode} interface is maintained as an attribute of \textit{RosPublisherScript}. When \textit{ChangeMode()} is invoked, the corresponding mode instance is assigned, and subsequent calls to \textit{Activate()} and \textit{Terminate()} are delegated to the active mode. This design enables different behaviors to be encapsulated behind a unified interface. Each mode also defines its own \textit{SendPose()} method, since the transmitted information and coordinate transformations differ across modes. For example, Follow Mode continuously sends target positions, Select Mode sends selected target positions intermittently, and Arm Mode continuously sends head-pose information. These mode-specific methods are called by \textit{RosPublisherScript::Update()}. To simplify implementation, we create separate game objects for each mode and attach the corresponding mode scripts.

\begin{figure}
    \centering
    \includegraphics[width=1\linewidth]{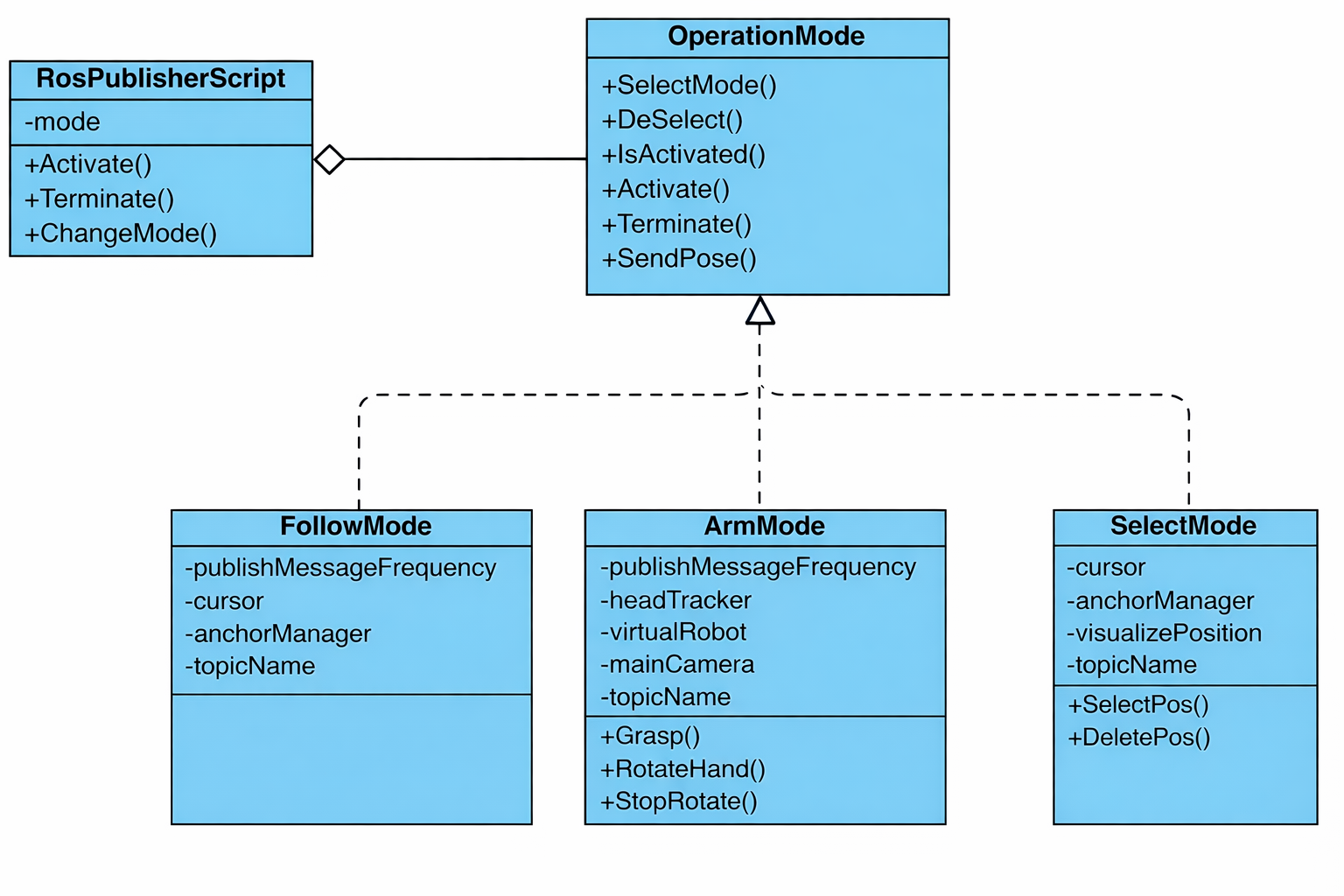}
    \caption{Class diagram for the state pattern used in mode switching. The \textit{RosPublisherScript} class holds an \textit{OperationMode} interface implemented by \textit{FollowMode}, \textit{SelectMode}, and \textit{ArmMode}. This design allows shared voice commands (\textit{Activate}, \textit{Terminate}) to invoke mode-specific behavior and enables each mode to define its own \textit{SendPose()} logic for different message formats and coordinate transformations.} 
    \label{fig:state_pattern}
\end{figure}

\textbf{Eye-gaze Tracking.} Since the application relies on eye gaze for robot locomotion control, accurate eye-gaze tracking is essential. The gaze ray is obtained through the HoloLens~2 eye-tracking API, and the eye cursor is placed at the intersection between this gaze ray and the world mesh reconstructed by the HoloLens~2. A challenge in this process arises from the cursor collider. Initially, in \textit{EyeGazeCursor::Update()}, we updated the cursor position every frame by moving it to the intersection point between the gaze ray and the mesh. However, the MRTK eye-gaze intersection interface~\cite{MRTK} considers all game objects in the scene, including the cursor itself. Once the cursor is moved to the detected intersection point, the gaze ray may intersect the cursor collider, causing MRTK to incorrectly treat the cursor as the new hit target. As a result, the cursor is pulled toward the camera instead of remaining at the intended location. To address this issue, we disable the box collider of the eye-gaze cursor. As shown in Fig.~\ref{fig:state_pattern}, the cursor is maintained as an attribute of both \textit{FollowMode} and \textit{SelectMode}, allowing these modes to generate control commands based on the cursor position.

\textbf{Head Tracking.} Head tracking is the most challenging part of the project since we need to handle many coordinate transformations. Since our goal is to let the robot arm mimic the behavior of the human head, a local coordinate of the robot hand in the robot frame needs to be specified.  As shown in Fig.~\ref{fig:head_motion}, we implement a head motion monitor and a virtual robot to extract the head motion and compute the local coordinate. We denote the transformation of the object $B$ under object $A$'s local frame as $T^A_B = (P^A_B, R^A_B)$, where $P$ and $R$ represent position and rotation respectively. Then we have:
\begin{equation}
\begin{aligned}
    T^{hand}_{robot} =& T^{head}_{v\_robot} = T^{head}_{world} T^{world}_{v\_robot} \\
    =& (T^{world}_{head})^{-1}  T^{world}_{v\_robot}.
\end{aligned}
 \label{eqn:transform}
\end{equation}
The initial hand position $T^{hand}_{robot}$ can be hard-coded as the offset of the real robot and the initial position of the virtual robot could be calculated as:
\begin{equation}
\begin{aligned}
    T^{world}_{v\_robot} = T^{world}_{head}  T^{hand}_{robot}.
\end{aligned}
\end{equation}
After initialization, once the user is moving, the head location $T^{world}_{head}$ can be directly assigned as the global coordinate of the camera, and we can update $T^{hand}_{robot}$ using Equation.~\ref{eqn:transform}. Instead of explicitly calculating the head position in the virtual robot's frame, we create a head tracker game object as a child of the virtual robot object, and we can directly get the transformation using \textit{headTracker.transform.localPosition}. 
\begin{figure}
    \centering
    \includegraphics[width=1\linewidth]{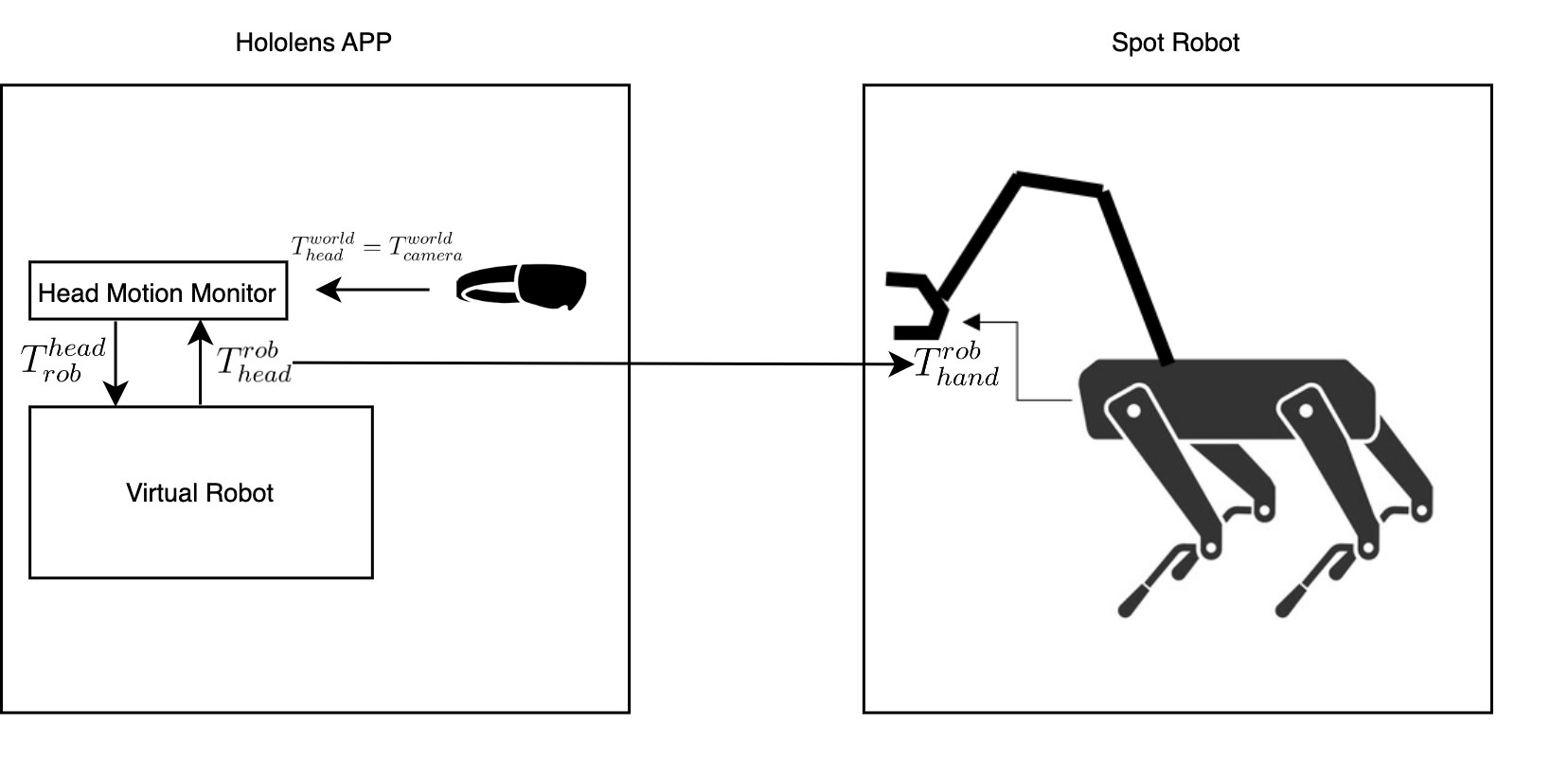}
    \caption{Coordinate transformation chain for head-to-arm mapping. The head pose in the world frame ($T^{world}_{head}$) and the virtual robot pose ($T^{world}_{v\_robot}$) are used to compute the hand pose in the robot frame ($T^{hand}_{robot}$). The virtual robot is a child of the head tracker in the Unity scene, enabling direct access to the relative transformation for real-time arm control.} 
    \label{fig:head_motion}
\end{figure}

An important issue for arms control is that the user cannot readily watch the target object when it is moving its head. To solve this problem, we store the local transformation of the virtual robot under the camera (head) frame and use this transformation to initialize the virtual robot's position when the arm control is activated again. Another issue is that the gripper angle may not be perfect for grasping items. Therefore, we enable the user to continuously rotate the gripper by tilting the head to adjust the angle. 

\textbf{Spatial Anchor.}\label{sec:anchor} 
The Azure Spatial Anchor can be used to co-localize Hololens and the Spot robot, in order to use it we used the Microsoft Azure Spatial Anchors package. When creating anchors, we first check if there are any existing anchors in the desired location, if there are not, we create a new anchor. Every time before sending positions, we transform the position into the anchor's local space. We could do this directly by calling \textit{anchor.transform.InverseTransformPoint()}. Because of the different coordinate systems used in Unity~\cite{unity} and Anchor~\cite{spatial_anchor}, we need to manually change the position we send from (x,y,z) to (z,-x,y). 

\textbf{User Interface.}

\textit{Voice Control.}
Voice commands provide simple and flexible ways to interact with the environment. To enable it in our application, we utilize the speech input system in MRTK~\cite{MRTK} together with the \textit{SpeechInputHandler} component. Different voice commands are specified in the \textit{MixedRealityToolkit object $>$ Input $>$ Speech} settings. Detailed response functions are set in the \textit{SpeechInputHandler} bounded to objects that handle the activities. For example, as the \textit{RosPublisher} object handles robot-related commands, one \textit{SpeechInputHandler} component is added there, and corresponding reacting functions are specified. To ensure that the voice recognition module works properly, a speech confirmation tooltip prefab is enabled. When a voice command is detected, a small box with the corresponding recognized command pops up in the view as shown in Fig.~\ref{fig:speech_tooltip}.
\begin{figure}
    \centering
    \includegraphics[width=1\linewidth]{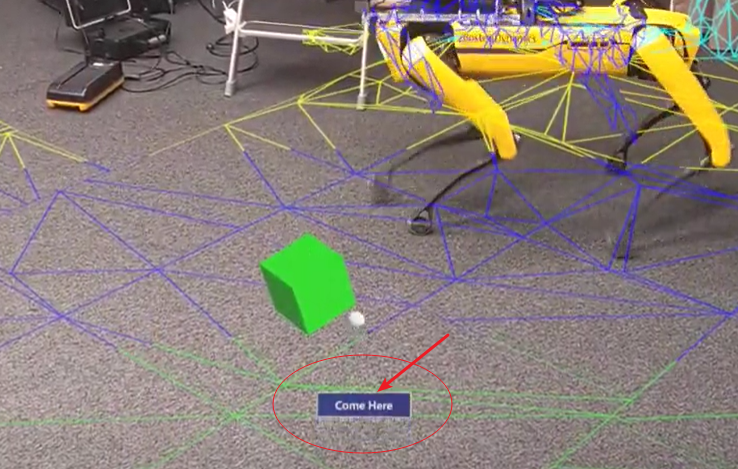}
    \caption{Speech confirmation tooltip displayed during voice interaction. When the user utters a command, a floating tooltip (indicated by the red arrow) shows the recognized phrase, providing immediate feedback and helping users with limited mobility verify that their intent was correctly captured.}
    \label{fig:speech_tooltip}
\end{figure}

\textit{Video Live Stream.}
When users operate the robot arm, sometime the view of the users would be blocked by the arm itself. To help the users have better views, a video live stream is added in Hololens, which is placed in an image plane, in front of the user, as shown in Fig.~\ref{fig:video_livestream}. The video is captured by the camera which is mounted on the gripper of the robot arm. But since the arm will rotate according to the head's motion, if we just simply fix the orientation of the image plane, the video itself would also rotate, which is hard for the user to watch. To avoid this problem, we subscribe to the orientation angle of the gripper by the ROS topic \textit{joint\_states}, and also apply this orientation change to the image plane, this way, even if the camera is rotated, the video is always adjusted to make sure to keep the right angle.

\begin{figure}
    \centering
    \includegraphics[width=1\linewidth]{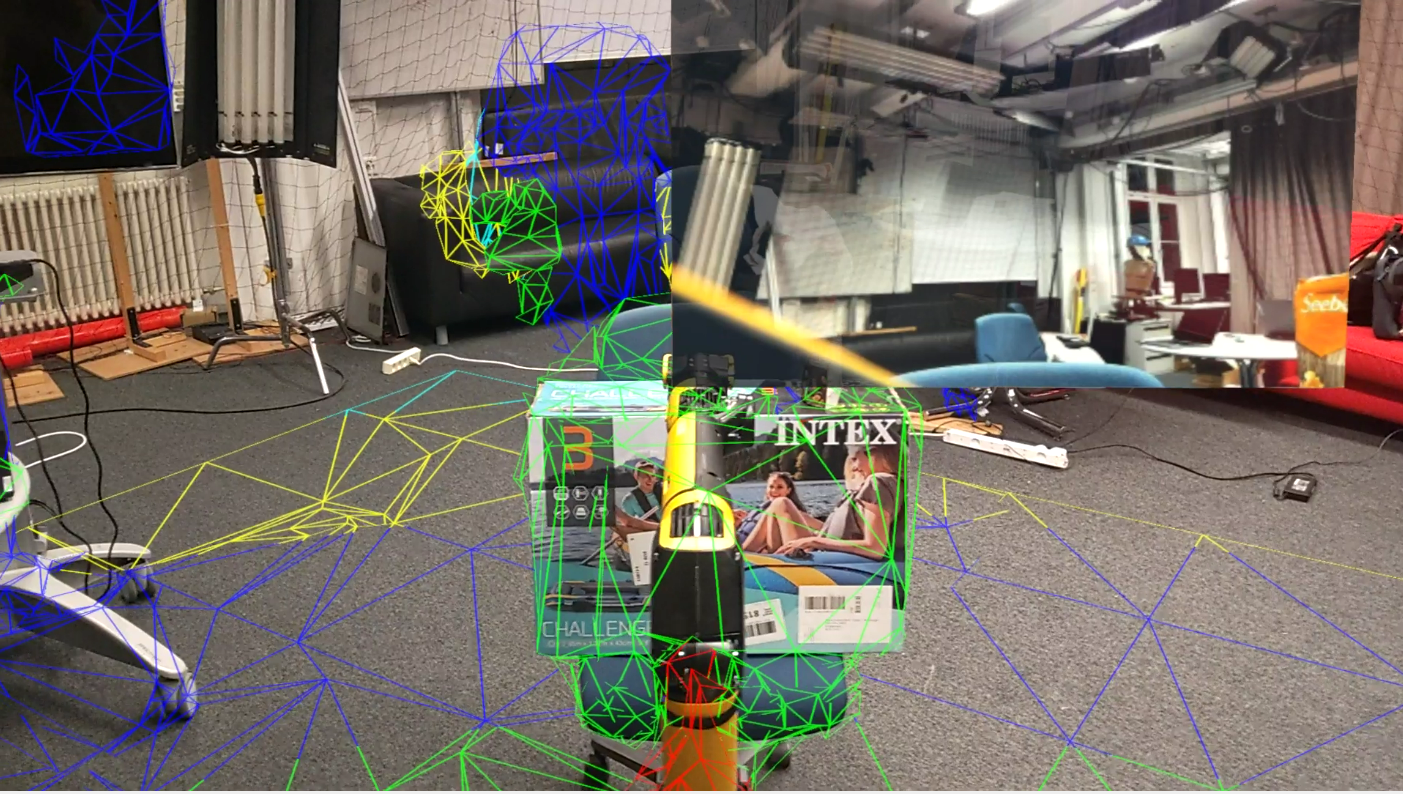}
    \caption{Gripper-camera video feed overlaid in the user's view. The live stream from the wrist-mounted camera is displayed in an image plane in the upper right corner. The plane orientation is updated via the \textit{joint\_states} topic so that the video remains upright and stable even as the gripper rotates with head motion.}
    \label{fig:video_livestream}
    \vspace{-1em}
\end{figure}

\textit{Help Panel.} 
The prefab for the help panel is from the MRTK Foundation package. On the panel, useful voice commands are listed as shown in Fig.~\ref{fig:help_panel}. The panel will pop up and disappear according to voice commands. Besides, it locates at position $(-0.5f, 0.25f, 2.5f)$ relative to the camera position whenever it is enabled. It does not change position according to user movement. It gives users necessary prompts when they interact with the Spot robot using HoloLens2.
\begin{figure}
    \centering
    \includegraphics[width=1\linewidth]{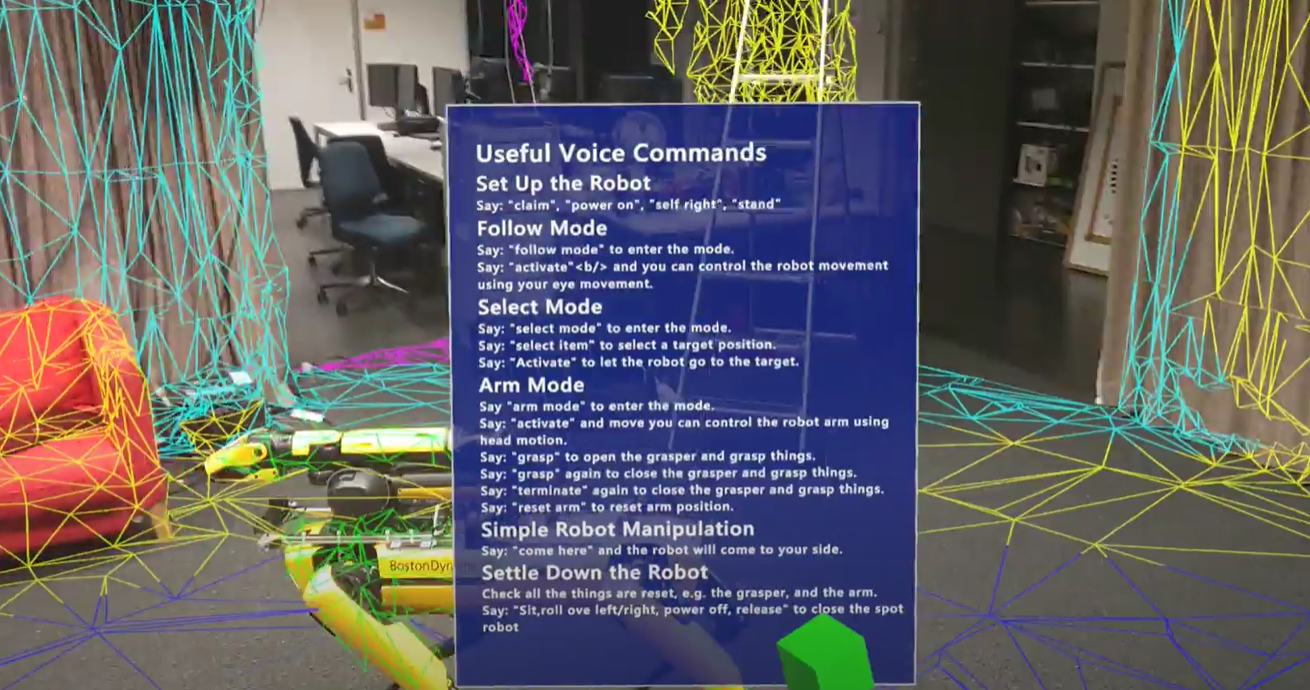}
    \caption{Help panel listing available voice commands. The panel is summoned by voice, positioned at a fixed offset from the camera, and provides quick reference for users when operating the Spot robot hands-free.}
    \label{fig:help_panel}
   
\end{figure}
 \vspace{-0.5em}
\subsection{Spot Robot and ROS} 
\textbf{Spatial Anchor Localization.} To co-localize Hololens and Spot robot, Spatial Anchor from Microsoft Azure~\cite{spatial_anchor} is used, as described in Sec.~\ref{sec:anchor}. The Spot robot needs to recognize the coordinate frame of the Spatial Anchor, which is achieved by the Spatial Anchor ROS package from Microsoft~\cite{anchor_ros}. Basically, we use the visual information collected by the camera of the Spot robot, and get the Spatial Anchor ID passed by the Hololens via ROS topic, then query the Anchor ID by Microsoft Azure. The coordinate frame of the certain Spatial Anchor is then added to the frame transformation tree of ROS.  

\textbf{Frame Transformation.} Since we have several coordinate frames (Unity~\cite{unity}, Spatial Anchor~\cite{spatial_anchor}, and ROS~\cite{ros}), all of them are represented in different coordinate systems, \ie Unity uses left-hand \textit{y}-up system, Spatial Anchor uses right-hand \textit{y}-up system, and ROS use right-hand \textit{z}-up system. To handle these different coordinate systems, we adjust the coordinate manually, by the ROS package \textit{spot-mr-core
}\cite{spot-mr-core} to transform all these three coordinate systems to the right-hand \textit{z}-up systems, before using the \textit{tf} package from ROS to do the frame transformation. This way, the destination coordinates sent by Hololens can be used directly in the ROS coordinate system.

\textbf{Spot Robot Movement.} For the robot movement, after we get the destination coordinate $(\triangle x, \triangle y)$ in the robot's body frame by the frame transformation, the robot will rotate $\theta$ angle along the $z$-axis to turn to the target direction and go to the target position simultaneously.  
\begin{equation}
\begin{aligned}
    \sin(\theta)&=\dfrac{\triangle y}{\left(\triangle y^2 + \triangle x^2\right)^{0.5}} \\ 
    \cos(\theta)&=\dfrac{\triangle x}{\left(\triangle y^2 + \triangle x^2\right)^{0.5}} \\ 
    \sin(\frac{\theta}{2}) &= \text{sign}(\sin(\theta))\sqrt{\frac{1-\cos{\theta}}{2}}\\
    \cos(\frac{\theta}{2}) &= \sqrt{1-\sin(\frac{\theta}{2})^2}\\
\end{aligned}
\end{equation}
The ROS topic \textit{/spot/go\_to\_pose} would be used to publish the desired pose in the robot's body frame: desired position $(\triangle x, \triangle y)$ and desired orientation $\text{Quaternion}(0,0,\sin(\frac{\theta}{2}),\cos(\frac{\theta}{2}))$.

\textbf{Spot Robot Driver.} We use the Spot Robot ROS Driver~\cite{spot_ros} to wrap the original Spot Robot Driver~\cite{spot_driver} into ROS. We use the ROS topic \textit{/spot/go\_to\_pose} to control the movement of the robot, the ROS service \textit{/spot/gripper\_pos} to control the pose of the robot arm, and the ROS service \textit{/spot/gripper\_angle\_open} to open/close the gripper. However, the original ROS service \textit{/spot/gripper\_pos} cannot adjust the operational time, it is fixed to 5 seconds to operate all the commands, which is too slow for our task, \ie our update rate is 0.5 seconds. To overcome this problem, we adjust the driver and pass one more parameter to describe how long to operate the command. This way, the robot arm can follow the user's command fluently.

\section{Experiment} 
\label{sec:user_study}
To evaluate the effectiveness of our product, we conduct a user study. In this stage, we assess users’ feelings when interacting with the application from usability, usefulness, and emotional aspects. The user experience is analyzed with quantitative and qualitative measurements. The experiment settings and user study results are described below.
\subsection{Experiment Settings}

Our goal is to test the two main functions of the application, namely using HoloLens2's eye tracking module to control robot walking and utilizing its head motion capture function for arm manipulation. The user study contains three parts, preparation, experiment conduct, and user feedback. In the preparation phase, we demonstrated to participants how to use the controller and Hololens 2 to control the robot. In addition, we gave participants 10 minutes per device to familiarize themselves with specific operations. In the experiment phase, we asked participants to complete the following two tasks with the controller and HoloLens 2, and recorded the time spent. The two scenarios are,
\begin{enumerate}
    \item Walking the robot from a specified starting point to a target location,
    \item Asking the user to touch a bottle on a table with the robot arm.
\end{enumerate}

Finally, we distributed questionnaires to participants and got their subjective evaluations of our application.

\noindent\textbf{Evaluation Metrics.}
Task performance and subjective ratings are considered quantitative metrics here. We use the task completion time to reflect task performance and it is an objective measurement. The participants’ ratings can give a highly-interpretational subjective reflection on their real feelings. Besides, qualitative assessments are included. We pay attention to the verbal feedback from users during experiments, and also ask them in the questionnaire about their psychological feelings, suggestions, and their thoughts on the accessible robot control topic.

\noindent\textbf{Questionnaire.}
The questionnaire contains 8 questions,
\begin{enumerate}
    \item Have you played with the Spot robot and/or HoloLens before? (1. Spot robot; 2. HoloLens; 3. None of them)
    \item How would you control the Spot robot if you have hand disabilities? (Open question)
    \item How would you rate your experience of robot movement using HoloLens follow mode? (Rate from 1 to 5; 1 means very bad, and 5 means very good)
    \item How would you rate your experience of robot movement using a controller? (Rate from 1 to 5; 1 means very bad, and 5 means very good)
    \item How would you rate your experience of robot arm movement using HoloLens arm mode? (Rate from 1 to 5; 1 means very bad, and 5 means very good)
    \item How would you rate your experience of robot arm movement using a controller? (Rate from 1 to 5; 1 means very bad, and 5 means very good)
    \item Do you think controlling via Hololens by our method is easier than the way you proposed? (1. Yes; 2. No)
    \item Any further suggestions for our application improvement? (Open question)
\end{enumerate}
They are distributed to users in \href{{https://docs.google.com/forms/d/e/1FAIpQLSeclS513aHQGXKppoGXtE0ieQa_XjvP6YQNdEDzD7Pr8pWoog/viewform}}{Google Form format}.
\subsection{Results}
\begin{table}
\setlength{\tabcolsep}{1.2mm}
  \centering
    \caption{User study results (n=11). \textbf{Time:} average completion time in seconds. \textbf{Score:} subjective rating on a 1--5 scale. Scenario~1: robot walking; Scenario~2: arm manipulation.}
  \begin{tabular}{@{}ccccc@{}}
    \toprule
    &\multicolumn{2}{c}{Average Time} & \multicolumn{2}{c}{Average Score} \\
    &Controller & HoloLens2 & Controller & HoloLens2 \\
    \midrule
    Senario 1 & 15.3s & 30.8s & 5.0/5.0 & 4.3/5.0 \\
    Senario 2 & 18.7s & 28.2s & 4.4/5.0 & 4.0/5.0 \\
    \bottomrule
  \end{tabular}
  \label{tab:user_study_quant}
  \vspace{-1.5em}
\end{table}

    
\begin{figure}[!ht]
    \centering
    \begin{minipage}[t]{0.48\linewidth}
        \centering
        \includegraphics[width=\linewidth]{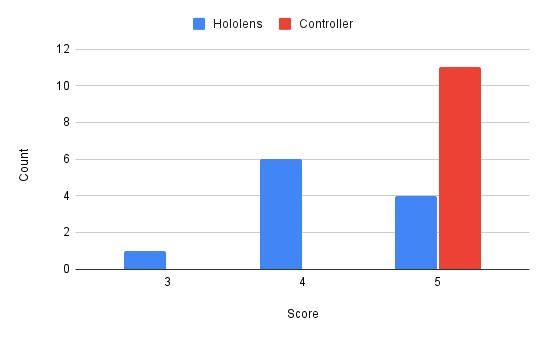}\\
        \small (a) Robot Locomotion
    \end{minipage}
    \hfill
    \begin{minipage}[t]{0.48\linewidth}
        \centering
        \includegraphics[width=\linewidth]{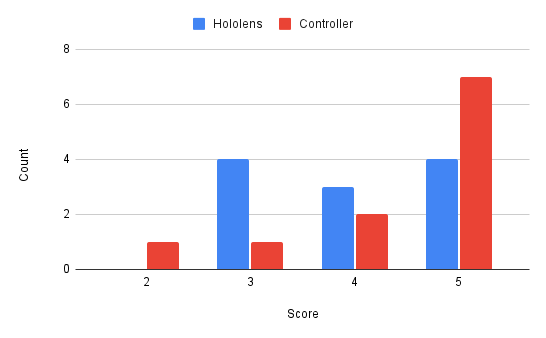}\\
        \small (b) Arm Manipulation
    \end{minipage}
    \caption{Distribution of subjective ratings on a 5-point scale for the two interaction scenarios: (a) robot locomotion and (b) robot arm manipulation. The results compare the handheld controller and HoloLens~2 in terms of user-perceived usability and satisfaction for each task.}
    \label{fig:robot_rate_dist}
    \vspace{-0.15in}
\end{figure}


So far, eleven users have taken part in our user study. Two of them have exposure to HoloLens, while the remaining nine participants have no previous experience with the two devices.
Their average task performance and subjective ratings are shown in Tab.~\ref{tab:user_study_quant} and detailed quantitative distributions are illustrated in Fig.~\ref{fig:robot_rate_dist}. Compared to the controller, users spend twice more time using HoloLens2 to move the robot to a specific location. Besides, it takes ten more seconds to operate the robot arm to touch the target item. Before making any conclusion, we need to clarify that our purpose is not to surpass the controller performance but take it as a baseline to reflect the smoothness and convenience of operating the robot using our developed application. Generally speaking, the controller provides a smoother experience, but the Hololens2 operation is also acceptable. Operating the robot arm using Hololens2 turns out to provide users with a comparable experience as a controller. The participants give us their solutions to accessible robot control in their responses to the questionnaire. Besides voice control and eye tracking we’ve applied in this project, they propose using body pose, foot movement, and EEG to operate the robot. 54.5\% (six persons) think their solutions perform similarly to ours and 45.5\% (five persons) consider our solution better. The users have provided us with valuable suggestions. For example, apply some safe collision avoidance strategies for the robot arm, and reduce the number of commands needed to switch among modes.
\section{Conclusion} 
\label{sec:future_work}
In conclusion, we present an egocentric mixed reality system for accessible multimodal control of the Spot robot using eye gaze, head movement, and voice commands. Our results highlight the promise of combining these modalities for intuitive and inclusive human--robot interaction. Future work will extend the system with new control modes, and vision--language--action models for more intelligent and context-aware behavior. We also aim to enable learning from human demonstrations and adaptation to user-specific behavior and preferences. Finally, broader user studies with more diverse participants are needed to better evaluate and refine the system.

\section{Acknowledgement}
We thank Boyang Sun for the support for the usage of the Spot robot from CVG lab. We also thank Dr. Iro Armeni for fruitful evaluation suggestions.

\end{document}